\pdfoutput=1

\documentclass[11pt]{article}

\usepackage{emnlp2022}

\usepackage{times}
\usepackage{latexsym}
\usepackage{amsmath}
\usepackage{array}

\usepackage[T1]{fontenc}

\usepackage[utf8]{inputenc}

\usepackage{microtype}
\usepackage{booktabs}
\usepackage{multirow}
\usepackage{graphicx}
\usepackage{amssymb}

\title{XDoc: Unified Pre-training for Cross-Format Document Understanding}

\author{Jingye Chen\thanks{Work done during internship at Microsoft Research Asia.}, Tengchao Lv, Lei Cui, Cha Zhang, Furu Wei\\ 
Microsoft Corporation\\
\texttt{\{v-jingyechen,tengchaolv,lecu,chazhang,fuwei\}@microsoft.com}
}

\begin{document}
\maketitle
\begin{abstract}
The surge of pre-training has witnessed the rapid development of document understanding recently. Pre-training and fine-tuning framework has been effectively used to tackle texts in various formats, including plain texts, document texts, and web texts. Despite achieving promising performance, existing pre-trained models usually target one specific document format at one time, making it difficult to combine knowledge from multiple document formats. To address this, we propose XDoc, a unified pre-trained model which deals with different document formats in a single model. For parameter efficiency, we share backbone parameters for different formats such as the word embedding layer and the Transformer layers. Meanwhile, we introduce adaptive layers with lightweight parameters to enhance the distinction across different formats. Experimental results have demonstrated that with only 36.7\% parameters, XDoc achieves comparable or even better performance on a variety of downstream tasks compared with the individual pre-trained models, which is cost effective for real-world deployment. The code and pre-trained models will be publicly available at \url{https://aka.ms/xdoc}.
\end{abstract}

\begin{figure}[t]
    \centering
    \includegraphics[width=0.48\textwidth]{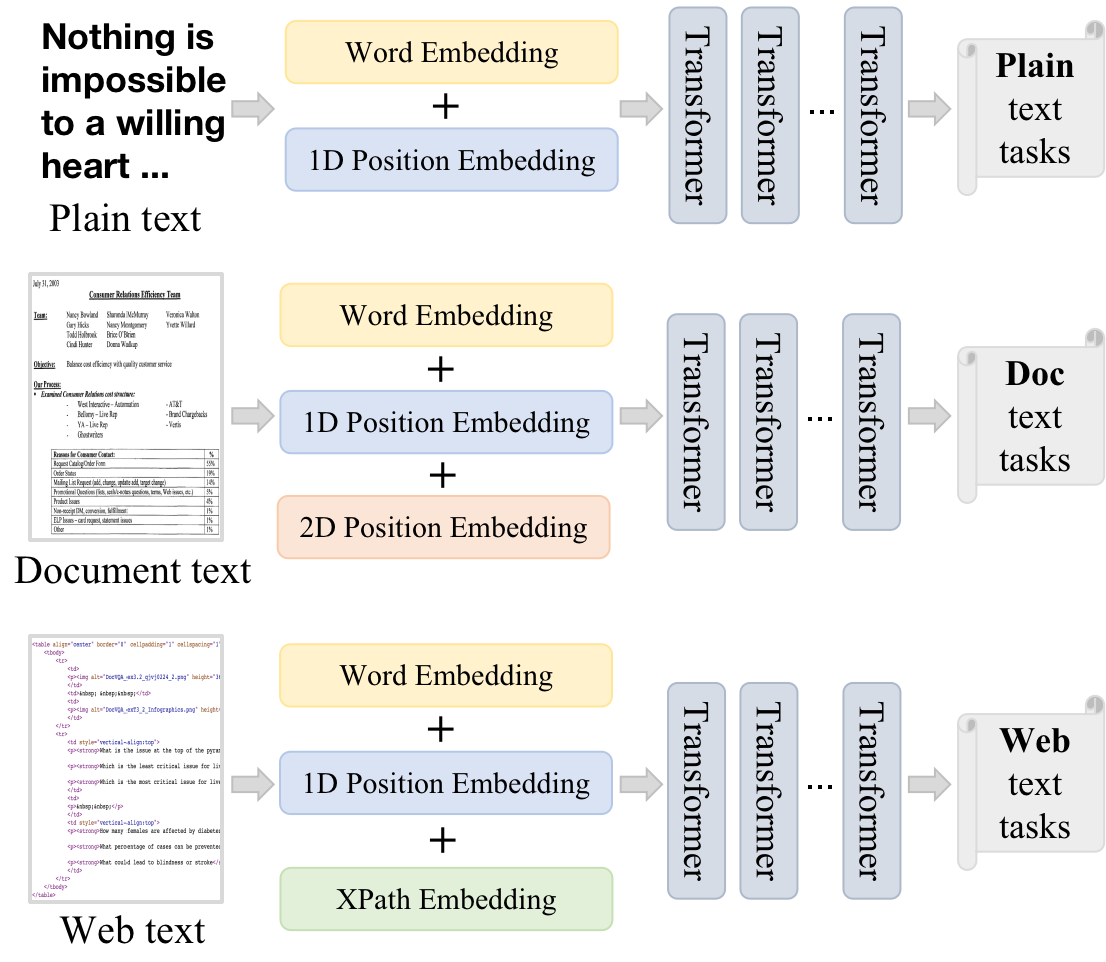}
    \caption{Pre-trained models for different document formats. Most of the structures are similar (word embedding, 1D position embedding, and Transformer layers) while only a small proportion of the structures (2D position and XPaths embedding) are different.}
    \label{fig:intro}
\end{figure}

\section{Introduction}
Document understanding has undoubtedly been an important research topic as documents play an essential role in message delivery in our daily lives \cite{cui2021document}. During the past several years, the flourishing blossom of deep learning has witnessed the rapid development of document understanding in various formats, ranging from plain texts \cite{devlin2018bert,liu2019roberta,dong2019unified}, document texts \cite{xu2020layoutlm,xu2020layoutlmv2,huang2022layoutlmv3}, and web texts \cite{,chen2021websrc,li2021markuplm,wang2022webformer}. Recently, pre-training techniques have been the de facto standard for document understanding, where the model is first pre-trained in a self-supervised manner (e.g. using masked language modeling as the pretext task \cite{devlin2018bert}) with large-scale corpus, then fine-tuned on a series of downstream tasks like question-answering \cite{rajpurkar2016squad,mathew2021docvqa}, key information extraction \cite{jaume2019funsd,xu2022xfund} and many others. Albeit achieving impressive performance on specific tasks, existing pre-trained models are far from flexible as they can only tackle texts in a single format (e.g. LayoutLM \cite{xu2020layoutlm} is designed for tackling document texts and is not suitable for web texts). 
This makes it difficult to combine knowledge from multiple document formats. Meanwhile, the category of pre-trained models will keep increasing if more formats (e.g. Word and PowerPoint) are further studied in academia.

\begin{figure*}[t]
    \centering
    \includegraphics[width=1\textwidth]{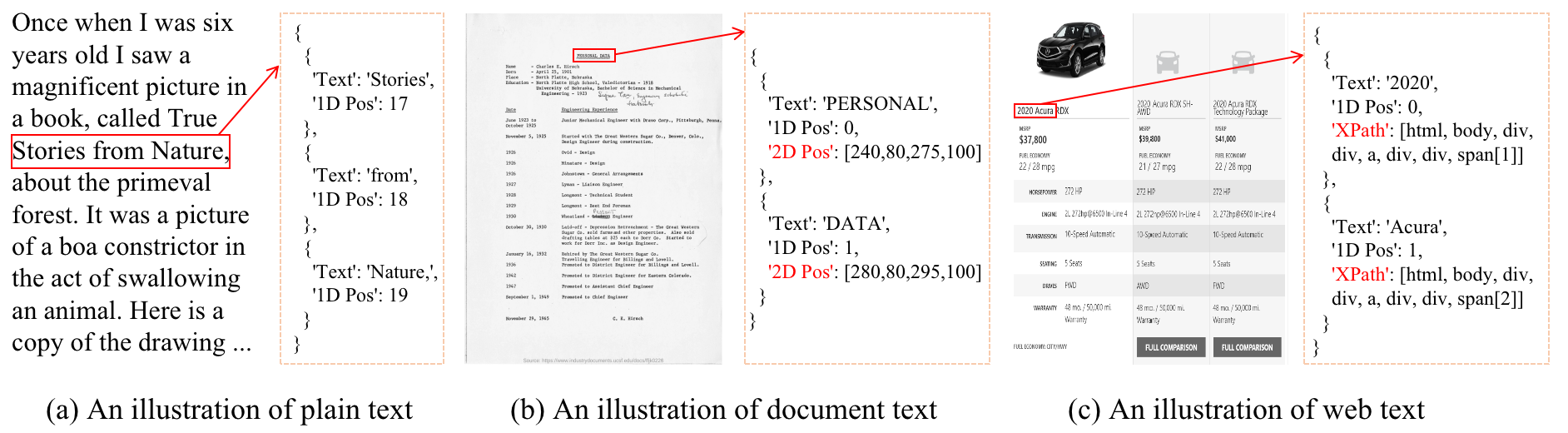}
    \caption{Illustrations of three document formats. For each format, the corresponding meta-information is shown in the dash boxes. Please note that the text content and 1D position are common attributes across three formats while 2D position and XPath strings (marked as red) are specific for document and web texts respectively.}
    \label{fig:illustration}
\end{figure*}

Among different pre-trained models for document understanding, it is observed that many pre-trained models share a similar architecture, such as a word embedding layer, a 1D position embedding layer, and Transformer layers (see Figure \ref{fig:intro}). In contrast, there are also different parts serving as prior knowledge for a specific format (e.g. two-dimensional coordinates for document texts and XPaths for web texts). Intuitively, we find that the parameters of different parts are far less than the parameters of the shared backbones. For instance, ${\rm LayoutLM_{BASE}}$ \cite{xu2020layoutlm} based on RoBERTa \cite{liu2019roberta} consists of 131M parameters while the 2D position embedding layer only contains 3M parameters (2.3\%). Similarly, ${\rm MarkupLM_{BASE}}$ \cite{li2021markuplm} based on RoBERTa has 138M parameters while the XPath embedding layer only contains 11M parameters (8.0\%). Therefore, it is indispensable to design a unified pre-trained model for various text formats while sharing backbone parameters to make models more compact. 

To this end, we propose XDoc, a unified architecture with multiple input heads designed for various categories of documents. For the sake of parameter efficiency, we share the backbone network architecture across different formats, including the word embedding layer, the 1D position embedding layer, and dense Transformer layers. Considering that the different parts only take up a small proportion in XDoc, we introduce adaptive layers to make the representation learning for different formats more robust. We collect the large-scale training samples for different document formats, and leverage masked language modeling to pre-train XDoc. Specifically, we use three widely-used document formats for experiments, including plain, document, and web texts (see Figure \ref{fig:illustration} for more details). To verify the model accuracy, we select the GLUE benchmark~\citep{wang2018glue} and SQuAD~\cite{rajpurkar2016squad,rajpurkar2018know} to evaluate plain text understanding, FUNSD~\cite{jaume2019funsd} and DocVQA~\cite{mathew2021docvqa} to evaluate document understanding, and WebSRC~\cite{chen2021websrc} for web text understanding. Experimental results have demonstrated that XDoc achieves comparable or even better performance on these tasks while maintaining the parameter efficacy.

The contributions of this paper are summarized as follows:

\begin{itemize}
    \item We propose XDoc, a unified pre-trained model that tackles texts in various formats in pursuit of parameter efficiency.
    \item Pre-trained with only masked language modeling task, XDoc achieves comparable or even better accuracy on various downstream tasks.
    \item The code and pre-trained models will be publicly available at \url{https://aka.ms/xdoc}.
\end{itemize}

\begin{figure*}[t]
    \centering
    \includegraphics[width=1\textwidth]{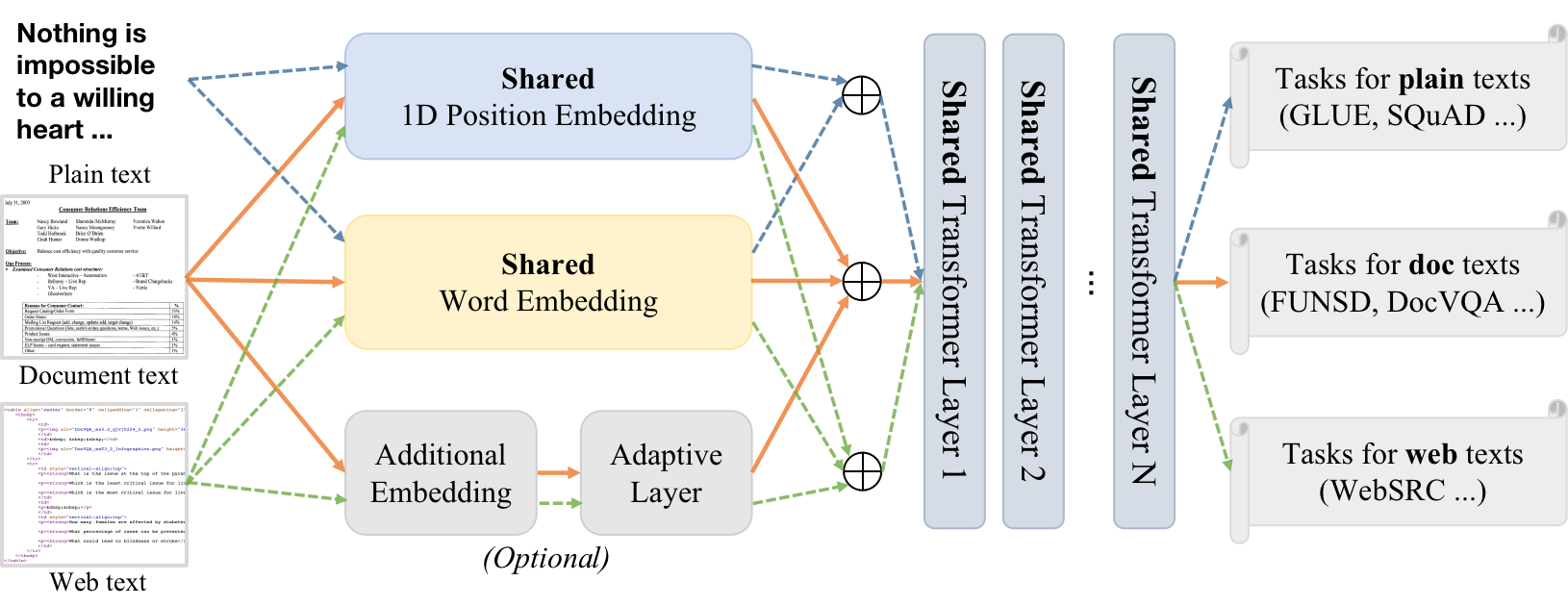}
    \caption{XDoc tackles multiple formats in one model while sharing most parameters including 1D position embedding, word embedding, and dense Transformer layers. An \textit{optional} embedding layer and adaptive layer are utilized for specific prior knowledge such as 2D position for document texts and XPaths for web texts (no additional prior for plain texts). We demonstrate the dataflow for document texts and use \textit{dash} lines for other formats.}
    \label{fig:architecture}
\end{figure*}

\section{XDoc}

In this section, we first introduce the architecture of XDoc and details of the embedding used for each document format, then introduce the objectives for pre-training the XDoc model.


\subsection{Model Architecture}
As is demonstrated in Figure \ref{fig:architecture}, XDoc is capable of tackling texts in various formats (plain, document, and web texts) in one model. For any input sequences, XDoc learns to embed them using a shared backbone and additional embedding layers when other prior knowledge is available. In detail, for any input text $T$, XDoc first tokenizes it into subwords $\mathbf{s}=s_{1:L}$ using WordPiece, where $L$ denotes the maximum length. Subsequently, for each subword $s_{i}$ with index $i$, it is first fed to a word embedding layer and we denote the output as ${\rm WordEmb}(s_{i})$. Then it is added with a learnable 1D position embedding ${\rm 1DEmb}(i)$. Since the word embedding and 1D position embedding layers are indispensable for Transformer-based models, we attempt to \textbf{share} the parameters across different formats. Based on this, we will detail the overall embedding for each document format in the next.

\paragraph{Overall embedding for plain texts} As there is no additional prior knowledge for plain texts, we simply add up the word embedding and 1D position embedding to construct the input for Transformer layers following \cite{devlin2018bert,liu2019roberta}. For each word $s_{i}^{P}$, where $i$ is the index and ``P'' denotes ``Plain’’, the overall embedding ${\rm Emb}(s_{i}^{P})$ can be calculated as follows:

\begin{equation}
{\rm Emb}(s_{i}^{P}) = {\rm WordEmb}(s_{i}^{P}) + {\rm 1DEmb}(i)
\label{eq:plain}
\end{equation}

\paragraph{Overall embedding for document texts} Different from the plain texts, the visually rich document texts are usually organized with 2-D layouts, where the coordinates of each text box play crucial roles in understanding. Hence, the 2D position should be necessarily taken into account during pre-training. Concretely, for a given subword $s_{i}^{D}$ (``D'' is the abbreviation of ``Document''), we denote the 2D position as $box_{i}^{D}=(l_{i}, r_{i}, t_{i}, b_{i}, w_{i}, h_{i})$, where $l, r, t, b, w, h$ denote left, right, top, and bottom coordinates, width and height of the text box, respectively. For example, as illustrated in Figure \ref{fig:illustration}(b), $l, r, t, b, w, h$ of the text ``PERSONAL'' is set to 240, 275, 80, 100, 35, and 20, respectively. Considering that most parameters are shared across different formats, we introduce an adaptive layer to enhance the distinction of specific prior information. The adaptive layer is simply implemented with a lightweight \textit{Linear-ReLU-Linear} sequence and we will discuss the effectiveness in Section \ref{sec:discussion}. Following \cite{xu2020layoutlm,xu2020layoutlmv2}, we add up all the embedding to construct the overall embedding ${\rm Emb}(s_{i}^{D})$ as follows:

\begin{equation}
\begin{aligned}
& {\rm Emb}(s_{i}^{D}) = {\rm WordEmb}(s_{i}^{D}) + {\rm 1DEmb}(i) \\ 
& \quad + {\rm DocAdaptive}[{\rm 2DEmb}(box_{i}^{D})]
\end{aligned}
\label{eq:doc}
\end{equation}

\begin{equation}
\begin{aligned}
& {\rm 2DEmb}(box_{i}^{D}) = {\rm LeftEmb}(l_{i}) + {\rm RightEmb}(r_{i}) \\ 
& \quad + {\rm TopEmb}(t_{i}) + {\rm BottomEmb}(b_{i}) \\
& \quad + {\rm WidthEmb}(w_{i}) + {\rm HeightEmb}(h_{i})
\end{aligned}
\label{eq:doc2}
\end{equation}
where ``${\rm LeftEmb}$'' denotes the embedding layer of the left coordinates (other embedding layers follow the same naming conventions). Please note that the  adaptive layer is not shared across different formats and ``${\rm DocAdaptive}$'' is specifically used for document texts.

\paragraph{Overall embedding for web texts} Since the 2-D layout of each website is not fixed and it highly depends on the resolution of rendering devices, we only employ XPath as the prior knowledge following \cite{li2021markuplm}. Concretely, for each subword $s_{i}^{W}$ (``W'' is the abbreviation of ``Web''), its XPath $xpath_{i}^{W}$ can be represented with a tag sequence and a subscript sequence. Taking the text ``Acura'' in Figure \ref{fig:illustration}(c) as an instance, its original XPath expression is /html/body/div/a/div/div/span[2]. Following MarkupLM \cite{li2021markuplm}, we construct the tag sequence as [html, body, div, a, div, div, span], representing the tag order from the root to the current node. In addition, the subscript sequence is set to [0, 0, 0, 0, 0, 0, 2], where each subscript denotes the index of a node when multiple nodes have the same tag name under a parent node (More explanations are shown in Appendix \ref{sec:xpath}). We add the tag embedding and subscript embedding to get the XPath embedding ${\rm XPathEmb}(xpath_{i}^{W})$. The overall embedding can be calculated as:


\begin{equation}
\begin{aligned}
& {\rm Emb}(s_{i}^{W}) = {\rm WordEmb}(s_{i}^{W}) + {\rm 1DEmb}(i) \\ 
& + {\rm WebAdaptive}[{\rm XPathEmb}(xpath_{i}^{W})]
\end{aligned}
\label{eq:web}
\end{equation}



Similarly, we leverage an adaptive layer ``WebAdaptive'' for better pre-training. Further, the overall embedding is fed to \textbf{shared} Transformer layers to obtain the contextual representations. 

\subsection{Pre-training Objectives}

We employ masked language modeling (MLM) as the pre-training task following \cite{devlin2018bert,liu2019roberta,xu2020layoutlm}. More specifically, we randomly mask 15\% of the input tokens, where 80\% tokens are converted to a special [MASK] token, 10\% tokens are randomly replaced with other tokens, and 10\% tokens remain unchanged. Through pre-training, the model needs to maximize the probability of the masked tokens according to the contextual representations.

\section{Experiments}
In this section, we first introduce the model configuration and detail the hyperparameters in XDoc, then introduce the pre-training strategies of XDoc. Next, we demonstrate the experimental results on a wide range of downstream tasks. At last, we verify the effectiveness of some designs in XDoc and have a discussion.

\subsection{Model Configurations}
The proposed XDoc is initialized with ${\rm RoBERTa_{BASE}}$, containing 12 Transformer layers, 768 hidden units, and 12 attention heads. The maximum length of each input sequence is set to 512 with a ${\rm [CLS]}$ token and a ${\rm [SEP]}$ token padding at the beginning and the end, respectively. The input sequence whose length exceeds 512 will be truncated, while the sequence shorter than 512 will be padded with ${\rm [PAD]}$ tokens.

\begin{table*}[ht]

\centering
\scalebox{0.8}{
\begin{tabular}{c c c c c c c c c c c c c}
\toprule
\multirow{2}*{\textbf{\#}} & \multirow{2}*{\textbf{Model}} & \multicolumn{3}{c}{\textbf{Pre-train}}  & \multicolumn{7}{c}{\textbf{Downstream Tasks}} \\
\cmidrule(lr){3-5}
\cmidrule(lr){6-10}
\cmidrule(lr){11-12}
\cmidrule(lr){13-13}
~ & ~ & P & D & W & MNLI-m & QNLI & SST-2 & MRPC & SQuAD1.1 / 2.0  & FUNSD & DocVQA & WebSRC  \\
\midrule
1 & ${\rm RoBERTa}$ & \checkmark & & & \textbf{87.6} & 92.8 & 94.8 & 90.2 & $\textbf{92.2}^{*}$ / $83.4^{*}$ & - & - & -  \\
2 & ${\rm LayoutLM}$ &  & \checkmark & & - & - & - & - & - & 79.3 & 69.2 & - \\
3 & ${\rm MarkupLM}$ & &  & \checkmark & - & - & - & - & - & - & -  & 74.5  \\
\midrule
4 & ${\rm XDoc_{100K}}$ &\checkmark  & &  & 87.0 & \textbf{93.0} & 95.2 & 90.1 & 91.9 / 83.4  & 70.1 & 64.5 & 58.5  \\
5 & ${\rm XDoc_{100K}}$ &  & \checkmark & & 86.7 & 91.3 & 94.5 & 89.9 & 91.4 / 82.9 & 87.3 & 69.4  & 58.6  \\
6 & ${\rm XDoc_{100K}}$ & &  &\checkmark & 86.5 & 92.0 & 94.6 & 90.1 & 91.4 / 83.1 & 71.6 & 63.6 & 64.8  \\
7 & ${\rm XDoc_{100K}}$ & \checkmark & \checkmark  & & 87.2 & 92.7 & 94.9 & 90.2 & 91.9 / \textbf{83.5} & 85.7 & 69.1 & 57.5  \\
8 & ${\rm XDoc_{100K}}$ &  & \checkmark & \checkmark & 86.4 & 91.6 & \textbf{95.3} & 91.0 & 91.7 / \textbf{83.5}  & 85.7 & 69.5 & 65.0 \\
9 & ${\rm XDoc_{100K}}$ & \checkmark  & & \checkmark & 86.8 & 92.3 & 95.1 & 90.6 & 91.6 / 83.0 & 70.0 & 64.7 & 64.8  \\
10 & ${\rm XDoc_{100K}}$ & \checkmark & \checkmark & \checkmark & 86.2 & 92.8 & 95.2 & \textbf{91.3} & 91.7 / 83.0 & 86.4 & 68.3 & 67.0   \\
\midrule

11 & ${\rm XDoc_{500K}}$ & \checkmark & \checkmark & \checkmark & 86.6 & 92.2 & 95.2 & 89.9  & 91.7 / 83.1 & 89.1 & 72.6 & 73.3 \\ 

12 & ${\rm XDoc_{1M}}$ & \checkmark & \checkmark  & \checkmark  & 86.8 & 92.3 & \textbf{95.3} & 91.1 & 92.0 / \textbf{83.5} & \textbf{89.4} & \textbf{72.7}  & \textbf{74.8} \\
\bottomrule
\end{tabular}}
\caption{Results on downstream tasks for various document formats. P, D, and W denote whether XDoc is pre-trained with plain, document, and web texts, respectively. Compared with methods designed for a specific format (\#1$\sim$\#3), XDoc achieves comparable or even better performance. \textbf{Accuracy} is used for MNLI-m, QNLI, and SST-2 for evaluation. \textbf{F1 score} is used for MRPC, SQuAD,  FUNSD, and WebSRC. \textbf{ANLS} is used for DocVQA. Digits marked with $^{*}$ denote that we re-implement the results since the original paper did not report them.}
\label{tab:experimental result}
\end{table*}

\subsection{Pre-training XDoc}
Large quantities of corpus play an essential role in learning robust representations during pre-training \cite{liu2019roberta}. Specifically, we utilize three categories of datasets for pre-training, which are detailed as follows.

\paragraph{Pre-training data for plain texts.} We follow \cite{liu2019roberta} to leverage five English-language corpora for pre-training, including B{\small OOK}C{\small ORPUS} \cite{zhu2015aligning}, English W{\small IKIPEDIA}\footnote{https://www.wikipedia.org/}, CC-N{\small EWS} \cite{ccnews}, O{\small PEN}W{\small EB}T{\small EXT} \cite{openweb}, and S{\small TORIES} \cite{trinh2018simple}, totaling 213,713 files for pre-training.

\paragraph{Pre-training data for document texts.} We leverage the large-scale scanned document image data IIT-CDIP Test Collection 1.0 \cite{lewis2006building} following \cite{xu2020layoutlm,xu2020layoutlmv2,huang2022layoutlmv3}. This dataset contains 42 million document pages, each of which is processed by OCR tools Tesseract\footnote{https://github.com/tesseract-ocr/tesseract} to yield the text contents and locations. For a fair comparison with previous works, we only use 11 million of them for pre-training.

\paragraph{Pre-training data for web texts.} Following MarkupLM \cite{li2021markuplm}, we take advantage of the large-scale dataset Common Crawl\footnote{https://commoncrawl.org/}, which contains petabytes of web pages in raw formats. Specifically, text contents and HTML tags are both available for each web page. According to \cite{li2021markuplm}, the authors first filtered Common Crawl with fastText \cite{bojanowski2017enriching} to remove non-English pages, then only kept common tags for saving disk storage, resulting in 24 million web pages for pre-training.

Specifically, we do not use any data augmentation or ensemble strategies for pre-training. We leverage AdamW optimizer \cite{loshchilov2017decoupled} with learning rate 5e-5 and epsilon 1e-8. Moreover, we linearly warm up in the first 5\% steps. Experiments are conducted with 32 NVIDIA Tesla V100 GPUs with 32GB memory. For those experiments pre-trained for 100K steps, we set the batch size to 128, while using all plain text datasets, the subset of document text (1 million), and web text (1 million) datasets for pre-training. Besides, we set the batch size to 512 and leverage all datasets for experiments pre-trained for 500K and 1M steps. FP16 is used during pre-training for accelerating and saving GPU memory. Within each batch, we equally sample documents in different formats for pre-training (see more 
discussions in Appendix \ref{sec:balance}).

\subsection{Fine-tuning on Downstream Tasks}
In this subsection, we utilize a wide range of downstream datasets to validate the ability of pre-trained XDoc in different formats. Specifically, for the plain texts, we leverage the widely-used GLUE benchmark \cite{wang2018glue} and SQuAD \cite{rajpurkar2016squad,rajpurkar2018know}. For document texts, we use the form understanding dataset FUNSD \cite{jaume2019funsd} and question-answering dataset DocVQA \cite{mathew2021docvqa}. For web texts, we utilize the question-answering dataset WebSRC \cite{chen2021websrc}.  
In the following, we will first introduce the downstream datasets in each format, then demonstrate the experimental results in detail.

\subsubsection{Fine-tuning on Tasks for Plain texts}
\paragraph{Fine-tuning on GLUE benchmark}
We evaluate XDoc on the General Language Understanding Evaluation (GLUE) benchmark \cite{wang2018glue}, which contains 9 datasets in total for evaluating natural language understanding systems. Specifically, 4 datasets four of them, including MNLI-m, QNLI, SST-2, and MRPC, are used for evaluation. We fine-tune XDoc for 10 epochs with a learning rate of 2e-5 and a batch size 16. The linear warmup is used for the first 100 steps. We utilize accuracy as the evaluation metric for MNLI-m, QNLI, SST-2, and F1 score for MRPC.

The experimental results are shown in Table \ref{tab:experimental result} and we leverage ${\rm RoBERTa_{BASE}}$ \cite{liu2019roberta} as the baseline (\#1). According to \#4, we notice that after pre-training with plain texts, the performance of XDoc is almost consistent with the baseline. It is intuitive since XDoc is initialized with ${\rm RoBERTa_{BASE}}$ and the continued training will not affect the performance. Interestingly, we notice that if XDoc is pre-trained without plain texts (refer to \#5, \#6, and \#8), the performance is still on par with the baseline, indicating that the knowledge of plain texts will not be easily forgotten when XDoc is pre-trained using other formats.

\paragraph{Fine-tuning on SQUAD V1.1 and V2.0}
We further employ the Stanford Question Answering Dataset (SQuAD)~\cite{rajpurkar2016squad,rajpurkar2018know} for evaluation. SQuAD contains two versions: SQuAD V1.1 and SQuAD V2.0. For V1.1, given a question, the answer can always be retrieved in the paragraph. By contrast, for V2.0, there are some questions that can not be answered, which is more challenging compared with V1.1. Specifically, XDoc is fine-tuned with 2 epochs for V1.1 and 4 epochs for V2.0. We set the batch size to 16 and the learning rate to 3e-5. We use the F1 score as the evaluation metric.

We also utilize ${\rm RoBERTa_{BASE}}$ \cite{liu2019roberta} as the baseline (\#1). As is demonstrated in Table \ref{tab:experimental result}, we notice that the performance does not fluctuate much under various pre-training settings (\#4$\sim$\#12). Similar to the experiment results on the GLUE benchmark, XDoc is capable of achieving comparable performance when pre-trained in all formats (refer to \#10$\sim$\#12). 


\subsubsection{Fine-tuning on Task for Document texts}

\paragraph{Fine-tuning on FUNSD}
We utilize the receipt understanding dataset FUNSD \cite{jaume2019funsd} to verify the ability of XDoc. Deriving from the RVL-CDIP dataset \cite{harley2015evaluation}, FUNSD contains 199 noisy scanned documents (149 samples for training and 50 for test) with 9,709 semantic entities and 31,485 words. Specifically, we focus on the entity labeling task, i.e. labeling ``question'', ``answer'', ``header'', or ``other'' in the given receipt. Concretely, we fine-tune XDoc for 1000 steps with the a batch size 64 and a learning rate 5e-5. We utilize linear warmup for the first 100 steps. The coordinates are normalized by the size of images following \cite{xu2020layoutlm}. F1 score is adopted as the evaluation metric.

For a fair comparison, we choose ${\rm LayoutLM_{BASE}}$ (\#2) \cite{xu2020layoutlm} as the baseline, which exploits the layout and text knowledge for tackling visually rich document understanding. Through the experimental results, we observe that XDoc can outperform the baseline by a large margin if document texts are used during pre-training. According to \#10, the performance can be boosted by 7.1\% if all formats are used for pre-training. Besides, we notice that the performance can be boosted further when XDoc is trained for more steps (further increase by 3.0\% according to \#12). In contrast, it is observed that the performance will heavily deteriorate if the document texts are absent during pre-training (decrease by 9.3\% according to \#9).

\paragraph{Fine-tuning on DocVQA}
For further validating the ability of XDoc on document texts, we utilize the document question-answering dataset DocVQA \cite{mathew2021docvqa}, which contains 10,194/1,286/1,287 images with 39,463/5,349/5,188 questions for training/validation/test sets, respectively. We follow LayoutLMv2 \cite{xu2020layoutlmv2} to employ Microsoft Read API to produce OCR results and find the given answers heuristically. We evaluate XDoc on the evaluation set and the final scores are obtained by submitting the results to the official website\footnote{https://rrc.cvc.uab.es/?ch=17}. We fine-tune XDoc for 10 epochs with a batch size 16 and a learning rate 2e-5. The linear warmup strategy is used for the first 10\% steps. Following \cite{xu2020layoutlm}, we normalize the coordinates by the size of images. We use Average Normalized Levenshtein Similarity (ANLS) as the evaluation metric.

As ${\rm LayoutLM_{BASE}}$ \cite{xu2020layoutlm} did not report the results on DocVQA, we borrow the ANLS score from LayoutLMv2 \cite{xu2020layoutlmv2}. Similar to the experimental results on FUNSD, we observe that the performance of XDoc highly depends on the participation of document texts during pre-training. For example, if XDoc is pre-trained without document texts, the performance drops by 4.7\%, 5.6\%, and 4.5\% according to \#4, \#6, and \#9. When pre-training with 100K steps using all formats, XDoc obtains comparable performance (refer to \#10). Furthermore, XDoc outperforms the baseline when training with more training steps (refer to \#11 and \#12).

\subsubsection{Fine-tuning on Task for Web Texts}
\paragraph{Fine-tuning on WebSRC}
We employ the Web-based Structural Reading Comprehension dataset (WebSRC) \cite{chen2021websrc} to verify the ability of XDoc on web texts. It contains 440K question-answer pairs collected from 6.5K web pages. The HTML source code, screenshots, and metadata are available in this dataset. The training/validation/test parts consist of 307,315/52,826/40,357 question-answer pairs. The answer is either a text span in the given web page or yes/no. We fine-tune XDoc for 5 epochs with a batch size 16, a learning rate 5e-5, and a linear warmup rate 0.1. F1 score is used as the metric. 

We use ${\rm MarkupLM_{BASE}}$ \cite{li2021markuplm} as the baseline (\#3). When XDoc is only pre-trained for 100K steps, we notice that the performance is subpar compared with the baseline. It is intuitive since MarkupLM is pre-trained with \textbf{three} pretext tasks, including masked language modeling, node relation prediction, and title-page matching. Interestingly, we observe that when training for more steps (\#12), the performance of XDoc surpasses the baseline. Similarly, it is observed that the performance will drop heavily if web texts are absent during pre-training (refer to \#4, \#5, and \#7).

\subsection{Discussions} \label{sec:discussion}
In this subsection, we conduct experiments to validate the effectiveness of the components or training strategies in XDoc. Unless specified otherwise, all experiments are pre-trained with 3M data (1M for each format) for 100K steps. Moreover, we discuss the parameter and time efficiency.

\begin{table}[t]
\centering
\scalebox{0.86}{
\begin{tabular}{c c c c c c}
\toprule
Init & MNLI-m & FUNSD & WebSRC & Avg \\
\midrule
Scratch & 75.4 & 78.8 & 29.2 & 61.1 \\
${\rm RoBERTa}$ & \textbf{86.2} & \textbf{86.4} & \textbf{57.5} & \textbf{76.7} \\
\bottomrule 
\end{tabular}}
\caption{Results on the initialization of XDoc.}
\label{tab:init}
\end{table}

\begin{table}[t]
\centering
\scalebox{0.9}{
\begin{tabular}{c c c c c}
\toprule
Layers & MNLI-m & FUNSD & WebSRC & Avg \\
\midrule
0 & 86.4 & 85.0 & 54.7 & 75.4 \\
1 & 86.2 & \textbf{86.4} & \textbf{57.5} & \textbf{76.7} \\
2 & \textbf{86.7} & 84.8 & 55.0 & 75.5 \\ 
3 & 86.4 & 86.1 & 55.7 & 76.1 \\
\midrule
$1^{\dag}$ & 86.4 & 84.8 & 57.3 & 76.2 \\
\bottomrule 
\end{tabular}}
\caption{Results on the symmetry and number of adaptive layers. $^{\dag}$ means that the document and web branches share the same adaptive layers.}
\label{tab:adaptive}
\end{table}

\paragraph{The initialization of XDoc}
We try to randomly initialize the parameters of XDoc with normal distribution and the results are demonstrated in Table \ref{tab:init}.  We observe that XDoc trained from scratch performs worse on downstream tasks, e.g. the performance drops by 10.8\% for MNLI-m, 7.6\% for FUNSD, and 28.3\% for WebSRC. Therefore, we choose to initialize XDoc with ${\rm {RoBERTa}_{BASE}}$ for better pre-training.

\paragraph{The symmetry and number of adaptive layers}
We utilize adaptive layers, which are implemented by a sequence of Linear and ReLU layers, to enhance the representations of different parts such as the 2D position and XPath embedding. Here we attempt to explore the symmetry and the number of adaptive layers. In detail, ``symmetry’’ means the document and web branches share the same adaptive layers. Additionally, we denote the number of layers as the number of ReLU layers (e.g. Layers=2 means \textit{Linear-ReLU-Linear-ReLU-Linear} and Layers=0 means no adaptive layers are used). As is demonstrated in Table \ref{tab:adaptive}, we notice that the average performance reaches the best if only one adaptive layer is used. Moreover, if we apply different adaptive layers to the document and web branches, the average performance can be boosted by 0.5\% compared with the counterpart (76.2\%).

\begin{table*}[t]
\centering
\scalebox{1.0}{
\begin{tabular}{c c c c c c c p{1.5cm}<{\centering}}
\toprule
\multirow{2}*{Methods} & Word & 1D Position & Transformer & 2D Position  & XPath & Adaptive & \multirow{2}*{\textbf{Total}} \\
~ & \small{39M} & \small{4M} & \small{85M} & \small{3M} & \small{11M} & \small{4M} & ~ \\
\midrule
${\rm RoBERTa}$ & \checkmark & \checkmark & \checkmark & - & - & - & 128M  \\
${\rm LayoutLM}$ & \checkmark & \checkmark & \checkmark & \checkmark & - & - & 131M  \\
${\rm MarkupLM}$ & \checkmark & \checkmark & \checkmark & - & \checkmark & - & 139M  \\
${\rm XDoc}$ & \checkmark & \checkmark &
\checkmark & \checkmark & \checkmark & \checkmark & 146M  \\
\bottomrule 
\end{tabular}}
\caption{Analysis of the parameter efficiency. XDoc shares most parameters across different formats, including word embedding, 1D position embedding, and Transformer layers. We omit some layers that contain negligible parameters such as segment embedding layers and LayerNorm layers. All the comparison models are in \textbf{base} size.}
\label{tab:parameters}
\end{table*}

\paragraph{Parameter efficiency}
We demonstrate some analysis of parameters in Table \ref{tab:parameters}. We observe that the word embedding and Transformer layers contain most of the parameters (124M), e.g. occupy 96.9\%, 94.7\%, and 89.2\% of all the parameters for ${\rm RoBERTa_{BASE}}$, ${\rm LayoutLM_{BASE}}$, and ${\rm MarkupLM_{BASE}}$, respectively. By sharing the word embedding, 1D position embedding, and Transformer layers across multiple text formats, the proposed XDoc is efficient in terms of parameter usage. In detail, the total amount of parameters is 398M for three single models, while XDoc only contains 146M parameters (146M/398M$\approx$36.7\%) but can be used for downstream tasks in multiple formats. Besides, the newly introduced adaptive layers only contain 4M parameters, which is almost negligible for the whole model (2.7\%).

\paragraph{Time efficiency} Apart from the newly introduced adaptive layer, the architecture of XDoc is similar to those models targeting one specific document format. Since the adaptive layer is lightweight, it will not take much time overhead. For example, when conducting inference on the DocVQA dataset, it costs 45 ms for a batch while the adaptive layer only consumes negligible 0.8 ms (1.8\%). Hence, XDoc is efficient in terms of the time cost. 

\section{Related Work}
In this section, we review the pre-trained methods for document understanding, ranging from plain, document, and web texts, respectively.

\paragraph{Pre-trained methods for plain texts} The understanding of plain texts through pre-training has been extensively studied during the last decade \cite{devlin2018bert,yang2019xlnet,bao2020unilmv2,liu2019roberta,lewis2019bart,lan2019albert,jiang2020convbert,he2020deberta,dong2019unified,lample2019cross,lin2021few,xue2020mt5}. For example, GPT \cite{radford2019language,brown2020language} utilizes Transformer \cite{vaswani2017attention} to conduct single-director masked-word prediction in an unsupervised manner. Besides, BERT \cite{devlin2018bert} utilizes two self-supervised tasks, including mask language modeling and next sentence prediction to obtain the robust representations of words based on Transformer. SpanBERT \cite{joshi2020spanbert} and ERNIE \cite{zhang2019ernie} try to mask consecutive text spans so as to construct a more challenging pre-train task. In \cite{dong2019unified}, the authors used different kinds of attention masks to enable one-direction and bi-direction attending. XLNet \cite{yang2019xlnet} introduces generalized 
autoregressive pre-training framework that utilizes a permutation language modeling objective. ELECTRA \cite{clark2020electra} first samples some candidates for the masked words and then uses a discriminator to predict whether a given token is replaced.

\paragraph{Pre-trained methods for document texts}
Benefiting from the public large-scale document dataset \cite{lewis2006building}, pre-training has become the de facto standard for analyzing document texts \cite{zhang2020trie,wang2021layoutreader,xu2021layoutxlm,li2022dit,appalaraju2021docformer,garncarek2021lambert,gu2022xylayoutlm,gu2022unified,wu2021lampret,wang2022lilt}. LayoutLM \cite{xu2020layoutlm} makes the first attempt to combine the Layout knowledge during pre-training to obtain robust contextual features for document texts. Based on LayoutLM, LayoutXLM \cite{xu2021layoutxlm} utilizes multilingual document text datasets for pre-training. StructuralLM \cite{li2021structurallm} jointly utilizes cell
and layout information from scanned documents to make the representations more robust.
LayoutLMv2 \cite{xu2020layoutlmv2} introduces a multi-modal architecture by combining additional image tokens in the Transformer. BROS \cite{hong2021bros} utilizes the token-masking and area-masking strategies for tackling information extraction tasks. XYLayoutLM \cite{gu2022xylayoutlm} proposes an Augmented XY-Cut algorithm to exploit proper reading orders during pre-training. Recently, LayoutLMv3 \cite{huang2022layoutlmv3} pre-trains the text branch and image branch simultaneously using Mask Language Modeling and Mask Image Modeling tasks, which makes it a robust model for tackling text-centric and image-centric tasks.

\paragraph{Pre-trained methods for web texts}
Compared with plain and document text analysis, the understanding of web texts is less studied and is more challenging since the layout of each website is not fixed (i.e. depending on the resolution of devices). MarkupLM \cite{li2021markuplm} takes the first attempt to incorporate web-based knowledge during pre-training while utilitzing three pretext tasks, including masked language modeling, node relation prediction, and title-page matching. Further, based on MarkupLM, DoM-LM \cite{deng2022dom} introduces a new pre-training task predicting masked HTML node. WebFormer \cite{wang2022webformer} simultaneously feeds text features and image features to the multi-modal Transformer while constructing rich attention patterns between these tokens.

Generally, although the mentioned methods show impressive performance in one specific format, they can not be transferred to tackle other formats. To mitigate this problem, the proposed XDoc is a scalable and flexible framework that is friendly to a wide range of formats, thus bringing much convenience for people.

\section{Conclusion and Future Work}
In this paper, we propose XDoc, a unified framework that can tackle multiple document formats (e.g. plain, document, and web texts) in one model. For parameter efficiency, XDoc shares most parameters, including the word embedding, 1D position embedding, and Transformer layers, across different document formats. The experimental results show that with only 36.7\% parameters, XDoc can achieve comparable or even better performance on downstream tasks spanning various document formats. For future work, we will consider exploiting the image features during pre-training to tackle image-centric tasks and designing more unified pre-training tasks for various document formats.

\section*{Limitations}
As XDoc only leverages the text and layout information for pre-training, it is not suitable to tackle some image-centric tasks such as page object detection. Besides, XDoc only uses masked language modeling as the only pre-training task in this version. For future work, we will consider designing more unified pre-training tasks for various document formats.

\bibliography{custom}
\bibliographystyle{acl_natbib}

\appendix

\section{Details of XPath embedding}
\label{sec:xpath}

\begin{figure}[ht]
    \centering
    \includegraphics[width=0.475\textwidth]{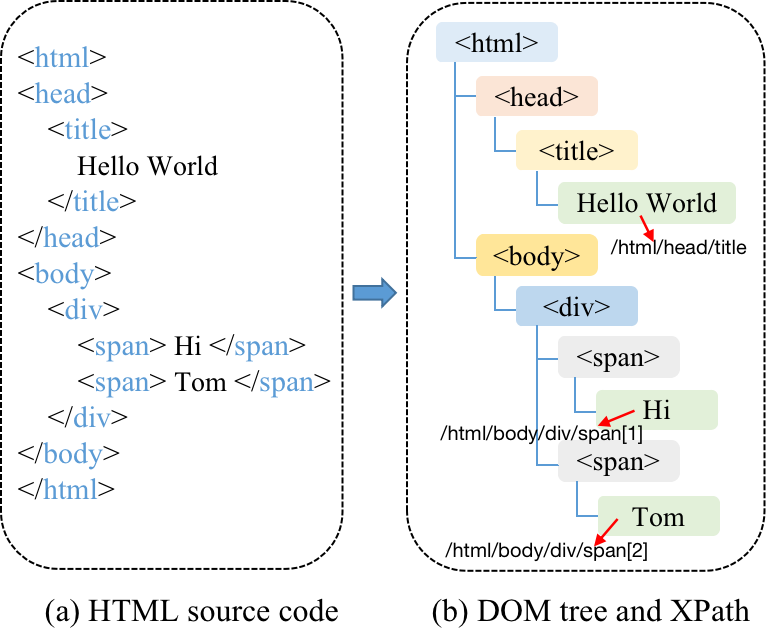}
    \caption{Illustrations of the way to construct XPath based on the corresponding HTML source code. Some examples of XPath are indicated using red arrows.}
    \label{fig:xpath}
\end{figure}

As is illustrated in Figure \ref{fig:xpath}, each web page can be represented as a DOM (Document Object Model) tree based on the corresponding HTML source code. In addition, XPath is a query language for selecting nodes based on the DOM tree. For example, the XPath of the text ``Tom’’ can be represented as “/html/body/div/span[2]”, where the texts denote the order of tag name traversed from the root node and the subscripts stand for the index of a node when more than one nodes have the same tag name under a parent node. For those tags without subscripts, we simply set the subscripts to 0. Following MarkupLM, we filter some unimportant tags and only reserve some common tags such as <html>, <body>, <div>, <span>, <li>, <a>, etc. 

To construct the XPath embedding for a given subword $s_{i}^{W}$, we first denote its XPath as $xpath_{i}^{W}$ = $[(tag_{1}, sub_{1}), (tag_{2}, sub_{2}), ..., (tag_{D}, sub_{D})]$, where $D$ means the maximum depth of the sequence, while $tag_{j}$ and $sub_{j}$ denotes the tag name and subscript at the $j$-th depth, respectively. For example, we represent the XPath of the text ``Tom'' as $[(html, 0), (body, 0), (div, 0), (span, 2)]$. Subsequently, for each pair $(tag_{j}, sub_{j})$ at depth $j$, we calculate its embedding $ts_{j}$ by adding up the tag embedding and subscript embedding:

\begin{equation}
\begin{aligned}
ts_{j} = {\rm TagEmb_{j}}(tag_{j}) + 
{\rm SubEmb_{j}}(sub_{j})
\end{aligned}
\label{eq:xpath1}
\end{equation}
Please note that the embedding layer of tags and subscripts vary across different depths. Finally, we concatenate the embedding of all pairs to construct the XPath embedding:

\begin{equation}
\begin{aligned}
& {\rm XPathEmb}(xpath_{i}^{W}) = 
[ts_{1}; ts_{2}; ...; ts_{D}]
\end{aligned}
\label{eq:xpath2}
\end{equation}

\section{Balance of Pre-training Data} 
\label{sec:balance}
We attempt to use different sampling ratios for different formats during pre-training and the experimental results are shown in Table \ref{tab:balance}. For example, ``3:1:1’’ denotes that there are approximately 60\% plain texts, 20\% document texts, and 20\% web texts in a batch. We notice that the average performance reaches the best (76.7\%) if we use the balanced sampling strategy. Interestingly, we observe that the sampling ratio with respect to one specific format does not positively correlate with the performance. For instance, when ``P:D:W’’ is set to 1:1:3, the performance on WebSRC is the worst (55.4\%) among all experiments.

\begin{table}[h]
\centering
\scalebox{0.9}{
\begin{tabular}{c c c c c}
\toprule
P:D:W & MNLI-m & FUNSD & WebSRC & Avg \\
\midrule
1:1:1 & 86.2 & \textbf{86.4} & \textbf{57.5} & \textbf{76.7} \\ 
3:1:1 & 86.7 & 83.8 & 56.7 & 75.7 \\
1:3:1 & 86.7 & 84.8 & 56.6 & 76.0 \\
1:1:3 & \textbf{87.1} & 83.7 & 55.4 & 75.4 \\
\bottomrule 
\end{tabular}}
\caption{Results on the balance of pre-training datasets. P:D:W denotes the ratio of plain, document, and web texts in a batch, respectively.}
\label{tab:balance}
\end{table}

\end{document}